# Dense Fusion Classmate Network for Land Cover Classification

Chao Tian
Harbin Institute of Technology
tianchao@sensetime.com

Cong Li
SenseTime Group Limited
licong@sensetime.com

Jianping Shi
SenseTime Group Limited
shijianping@sensetime.com

## Abstract

*Recently, FCNs based methods have made great progress in semantic segmentation. Different with ordinary scenes, satellite image owns specific characteristics, which elements always extend to large scope and no regular or clear boundaries. Therefore, effective mid-level structure information extremely missing, precise pixel-level classification becomes tough issues. In this paper, a Dense Fusion Classmate Network (DFCNet) is proposed to adopt in land cover classification. DFCNet is jointly trained with auxiliary road dataset seemed as "classmate", which properly compensates the lack of mid-level information. Meanwhile, a dense fusion module is also integrated, which guarantees the precise discrimination of confused pixels and benefits the network optimization from scratch. Score on Deep-Globe land cover classification competition shows that our approach has achieved good performance.*

## 1. Introduction

In recent years, convolution neural network (CNN) based models have achieved huge success in a wide range of tasks of computer version, such as semantic segmentation, which has a wide array of applications for scene understanding. The problem of land cover classification in 2018 Deep-Globe CVPR Satellite Challenge [6] can be seen as a multi-class semantic segmentation task. Many works have outperformed of image segmentation, e.g. [5, 2, 9, 17, 16, 11]. Semantic segmentation requires to make predictions at every pixel. There are three main categories of methods to improve semantic segmentation, encoder-decoder architecture, feature fusion and strengthing the spatial information. The encoder-decoder architecture has been widely used in recent semantic segmentation models, such as U-net [14], Deconvolutional network [20] and SegNet [1].They all recover the input spatial resolution at their outputs with an upsampling path. Deconvolution network and SegNet use the upsampling and max-pooling indices with the stack of simple convolution layers. The dilated convolutions [18], instead of max-pooling, was used in the backend of CNN

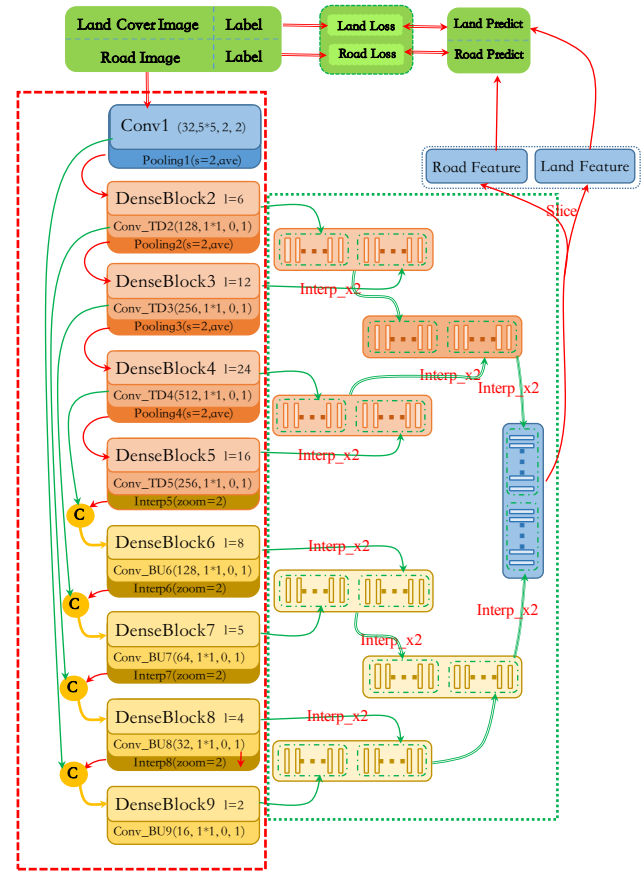

Figure 1. Architecture of our DFCNet

models in Deeplab [3, 4] to generate high resolution coarse score maps.

From global to local, semantic segmentation needs rich representations that span levels from low to high for more information aggregation. Some works have focused on the feature fusion, such as [22, 19] . The spatial pyramid pooling [8] (SSP) generates a fixed-length representation regardless of image size/scale, which helps to merge multi scale information to get more information for segmentation at a deeper stage of the network hierarchy. The pyramid scene parsing network [21] (PSPNet) exploit the capability of global context information by using pyramid pooling on the feature map of the ResNet [7].

Spatial information can enhance the semantic information of the network. In order to get more spatial information, Many works aim at changing the architecture of the network to better take use of the spatial structure, such as Spatial CNN (SCNN) [13]. SCNN transforms the traditional concatenation of layer-by-layer connections into slice-by-slice conjugations in feature maps, enabling the pixel rows and columns in the graph to pass messages. The Spatial Propagation network [12] learns affinity by constructing a row/column linear propagation model. In addition, there are still some works trying to better understand the scene information and increase the generalization of the network. Multi task learning [15] (MTL) aims at simultaneous training of multiple tasks with multiple data sets.

Since remote sensing scenes have their own characteristics, general semantic segmentation networks may not be totally suitable. For example, most of the geographical elements have exaggerated scales, such as water, forest, agriculture and so on, so richer and larger scope context information is necessary for precise result predicting. What's worse, without regular geometry shape, general effective structural information is so lack that more accurate details are desired. Fortunately, we found that roads have distinguishable distribution on different classes of land cover, and roads dataset is offered by another DeepGlobe challenge. We hypothesis that road is a property of land cover, implicitly containing effective structure information. So, considered as an auxiliary class for land cover classes, road dataset is jointly used to address land cover classification tasks. Inspired by approaches mentioned above and further exploration of satellite image specific properties, a novel architecture is presented in this paper, which we call Dense Fusion Classmate Network (DFCNet). The contributions of our method can be summarized as:

1. Propose a classmate strategy like multi-task learning, which successfully combines two seemingly unrelated tasks dataset and obtains obvious improvement on specified task.

2. Adopt a dense fusion module, results to advantages of gradient flow, feature refinement and multi-scale fusion, dense supervision.

3. Provide a meaningful perspective that road class is an strong compensate for other land cover classes, road distribution can be viewed as a kind of structure to help to distinguish confused land cover.

## 2. Approach

To better understanding our DFCNet, we firstly introduce the whole architecture, then introduce the Classmate strategy and Dense Fusion module from the details.

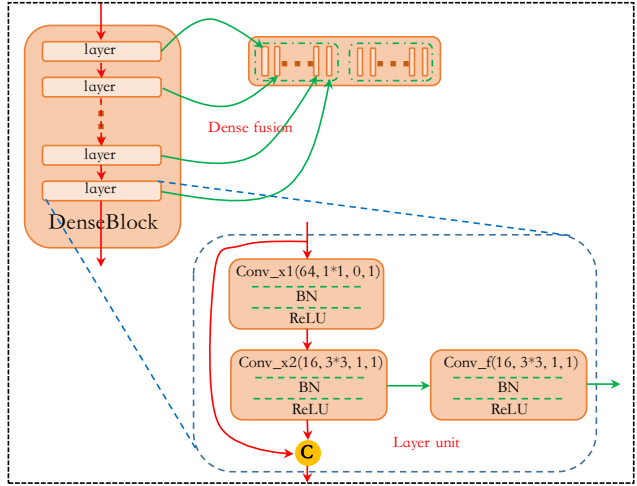

Figure 2. Detailed illustration of Dense Block and the Dense Fusion

### 2.1. Architecture

We construct our Dense Fusion Classmate Network (DFCNet) based on the FC-DenseNet, which uses the average-pooling for downsampling and an interp layer using the bilinear interpolation for upsampling.The overall architecture of DFCNet is illustrated in Figure 1. We just show the groups of dense block 2 to 9, and ignores the detail branches of conv2-x to conv9-x since it is too deep. To better explain our network, we take the DenseBlock 5 for example, l = 16 means that there are 16 layer units in this block. (256, 1*1, 0, 1) means that the channel number is 256, the kennel size is 1*1, the padding equals 0 and stride equals 1. In the Dense Fusion module, two neighouring dense blocks integrated together, but there are no overlap between every two blocks, and our Dense Fusion recursively integrating all dense blocks to the final level. Detailed dense block with Dense Fusion module is illustrated in Figure 2. All convolution layers in dense blocks are integrated, and the lower resolution blocks are upsampled before integration. The integration is implemented by an element-wise sum. The training data from road and land is merged by a concat layer at the first dimensionality.

### 2.2. ClassmateNet

How to better understanding the scene is very critical, since the unique characteristic of remote scene images. Trough the analysis of data, we find that data labeling also brings some noise to our training. The dividing line between the rangeland an forest is not particularly clear, which is also influenced by the characteristics of remote sensing data itself. By the visual analysis of images, we discover that some road appears on the map, but it dose not belong to our semantic segmentation classes. We further find that some of more obvious roads tend to appear between urban, agriculture and rangeland and intensity roads are more likely to appear in urban. This is an important information for our classification, which helps us to better distinguish these classes with data annotation ambiguities in scene semantic segmentation. Based on this information, we proposed the Classmate strategy in our DFCNet, which uses both road and land data for training and let the road information learned from the net to help the land semantic segmentation. The training data from the road extraction in 2018 DeepGlobe CVPR Satellite Challenge has similar scene to the land cover. We take it as ancillary data for network training by helping generate road information. Land segmentation is unstructured because it does not contain explicit boundary information. However, The road can help our ClassmateNet to learn the structure information in the mid-level, which just make up for the lack of land cover information.

The ClassmateNet gets score of 51.87 on the valid dataset, which is 8.3 points higher than the baseline. We save the feature map of the deepest convolution layer of both our baseline and ClassmateNet with the same downsample rate. As shown in Figure 3, we choose three input images from the test dataset, and visualize their feature maps. As we can see from feature maps, intensive roads tend to appear in urban, which makes the segmentation of the urban more accurate. Information from roads make the boundaries of segmentation classes more smoother, which can give us less ambiguity and uncertainty in segmentation. The visualization result of feature maps is consistent with our score on the valid dataset.

### 2.3. Dense Fusion

Dense Fusion integrated both shallow and deep, which makes the prediction on the level that contains the whole information from global and local and resolution from fine to coarse. Also the loss can be quickly back propagation to both deep and shallow layer, which makes the network better supervision.

Our DFCNet get score of 52.24 in the valid dataset. Although we did not get a great improvement in scores, but the details were handled better in terms of visual effects. By visualization result of the valid images, we find that the Dense Fusion can make the segmentation more meticulous, since the information merged from the shallow and deep layer can make the prediction get both global and local information, then a better performance.

## 3. Experiment

After the introduction of the approaches, we provide a brief description of our dataset, implement details and the results on the valid and test dataset.

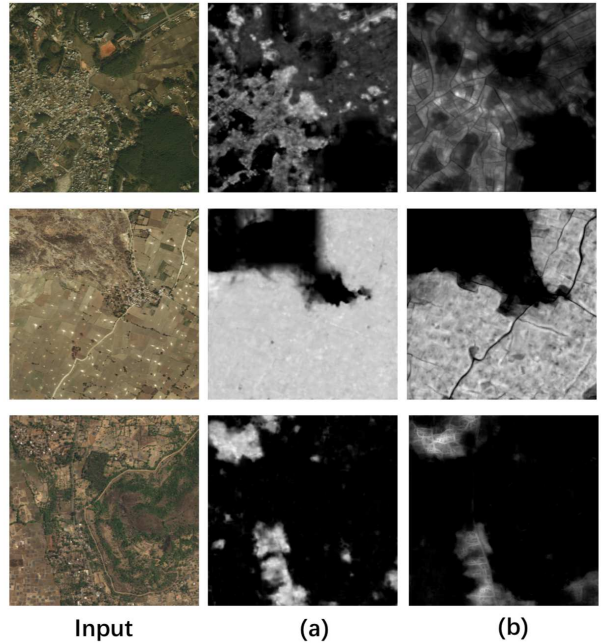

Figure 3. Feature maps of (a) our baseline and (b) DFCNet from the deepest layer with the same downsample rate.

### 3.1. Dataset

We use the dataset from the DeepGlobe Land Cover Classification and Road Extraction to build our model. 5000 images of road training dataset, each of resolution 1024 x 1024, and 600 images of land training dataset, each of resolution 2448 x 2448, are used for our training. 203 images of land dataset are used for our testing.

### 3.2. Implementation Details

We use caffe [10] as the deep learning framework. All of our models take the FC-DenseNet as the backbone. Based on the standard DenseNet 121, we set hyper-parameters following existing DenseNet work. We train on 8 GPUs (effective mini-batch size is 32) for 80k iteration, with a learning rate of 0.001, and use a weight decay of 0.0005, momentum of 0.9 and the ploy optimizer strategy. We did a series experiments on different crop size from 513 to 1025 and get different results on the test dataset sliced by ourself from the training dataset, including 203 images. It shows that the relatively bigger crop size can get a better performance, and we finally choose crop size of 1025 x 1025. For the FC-DenseNet we trained, we make the prediction on the downsampling of 4 times and get score of 43.57 on the valid dataset as our baseline. We set the hyper-parameter of DFCNet the same as the FC-DenseNet but take the DenseNet model as initial model.

As for our DFCNet, we take the road and land dataset for training. We set the crop size of 1025 and random re-size varying in the range of [0.5,1.5], [0.8,1.25] for road and land data. The effective mini-batch size is 16 due to the limitation of GPUs. Two datas are merged by a concat layer, and sliced in the deepest convolution layer. To learn the unique feature of this two tasks, we add different convolution layers after the slice layer, and make the prediction on the downsampling of 2 times.

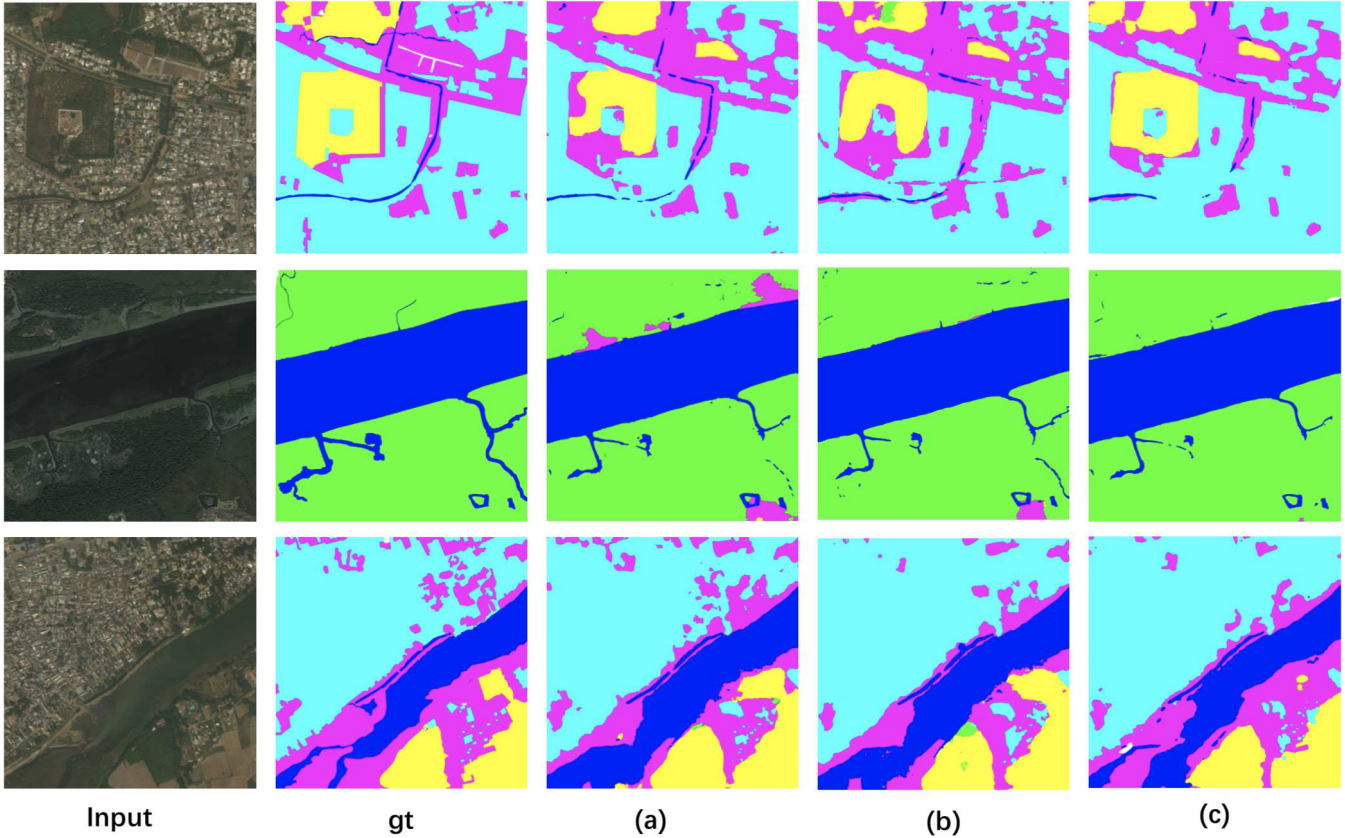

Figure 4. Result visualization comparisons: ground truth, (a) Baseline, (b) Classmate, (c) DFCNet.

### 3.3. Results

The prediction mean intersection over union (mIOU) of the land cover classification is presented in Table 1. Training is done on the trainval set and testing on the valid dataset. And our DFCNet gets score of 54.13 on the test dataset. In order to reduce randomness of the prediction process and reduce the noise caused by the data itself, we make the prediction on multi scales and fusion all scales prediction to the final one, and we also merged some models based on our DFCNet to predict on the test. By doing this, we get score of 55.59.

We choose some visualization results of the test dataset sliced by ourself from the training data, as illustrated in Figure 4. As we can see, our DFCNet can do better for details than baseline.

Table 1. *Prediction score (in%)* of different methods on valid dataset of the land cover classification task.

| Methods | Baseline | ClassmateNet | DFCNet |
|---------|----------|--------------|--------|
| mIOU    | 43.57    | 51.87        | 52.24  |

## 4. Conclusion

This paper presented a novel method for land cover classification, namely Dense Fusion Classmate Network (DFCNet). DFCNet is inspired by FC-DenseNet [11], but has two uniqueness. First, a classmate strategy is introduced, which successfully combines two seemingly unrelated tasks dataset and provides rich mid-level structural information. Second, a dense fusion module is integrated into DFCNet, with advantages: gradient flow, feature refinement and multi-scale fusion, dense supervision. Finally, competitive score on the DeepGlobe land cover classification challenge has demonstrated the potential of DFCNet, without any extra dataset or pre-train model.

195